\def\year{2021}\relax
\documentclass[letterpaper]{article} 
\usepackage{aaai21}  
\usepackage{times}  
\usepackage{helvet} 
\usepackage{courier}  
\usepackage[hyphens]{url}  
\usepackage{graphicx} 
\usepackage{layouts}
\usepackage{floatrow}
\usepackage{subfig}
\usepackage{amsmath}
\usepackage{amsfonts}

\usepackage{natbib}  
\usepackage{caption} 
\title{Task Uncertainty Loss Reduce Negative Transfer in Asymmetric Multi-task Feature Learning}
\author{

    Rafael Peres da Silva,\textsuperscript{\rm 1,2}
    Chayaporn Suphavilai,\textsuperscript{\rm 2} 
    Niranjan Nagarajan\textsuperscript{\rm 1,2,3}\\
}
\affiliations{

    \textsuperscript{\rm1}Department of Computer Science, School of Computing, National University of Singapore \\
    \textsuperscript{\rm2}Laboratory of Metagenomic Technologies and Microbial Systems, Genome Institute of Singapore, A*STAR \\
    \textsuperscript{\rm 3}Yong Loo Lin School of Medicine, National University of Singapore \\
    rafael@comp.nus.edu.sg, suphavilai\_chayaporn@gis.a-star.edu.sg, nagarajann@gis.a-star.edu.sg

}

\begin{document}

\maketitle

\begin{abstract}
Multi-task learning (MTL) is frequently used in settings where a target task has to be learnt based on limited training data, but knowledge can be leveraged from related auxiliary tasks. While MTL can improve task performance overall relative to single-task learning (STL), these improvements can hide negative transfer (NT), where STL may deliver better performance for many individual tasks. Asymmetric multi-task feature learning (AMTFL) is an approach that tries to address this by allowing tasks with higher loss values to have smaller influence on feature representations for learning other tasks. Task loss values do not necessarily indicate reliability of models for a specific task. We present examples of NT in two orthogonal datasets (image recognition and pharmacogenomics) and tackle this challenge by using aleatoric homoscedastic uncertainty to capture the relative confidence between tasks, and set weights for task loss. Our results show that this approach reduces NT providing a new approach to enable robust MTL.
\end{abstract}

\section{Introduction}
\noindent 
In the presence of task-relatedness, multi-task learning (MTL) jointly trains all tasks by leveraging common feature patterns, allowing for a shared representation as well as improvements in overall predictive performance. By borrowing information learnt from related tasks, MTL approaches can offer better results for tasks with small sample sizes. This scenario is often the case in pharmacogenomics and bioinformatics where patient samples can be limited and omics assays can be expensive. Despite the overall benefits of MTL, negative transfer (NT) behavior can be a limitation where task-specific predictive performance based on MTL can be inferior to a corresponding single-task model (STL). While there are potentially many reasons for observing NT, the impact of a subset of unrelated tasks on the multi-task learning process is a common concern in bioinformatics. Assigning equal weights to all tasks in MTL can result in poorer performance for unrelated tasks, as well as for related tasks where the contributions to the shared representation may vary. To overcome this, Lee \cite{Lee1} proposed asymmetric knowledge transfer learning (AMTL) by allowing a larger contribution from reliable low-loss tasks while reducing the contribution of unreliable related tasks. Subsequently, Lee \cite{Lee2} extended the method to enable asymmetric task-to-feature transfer by reconstructing shared parameters with an auto-encoder regularization guided by low-loss tasks. However, task loss is not a good indicator of reliability as low-loss might result from overfitting or lack of generalization, especially in small sample size settings. To represent task reliability, task-uncertainty \cite{Kendall} can provide a better representation as it is less prone to low-loss misleadings. We extend the state-of-the-art Deep Asymmetric Multitask Feature Learning (Deep-AMTFL), which relies on task-loss, by incorporating aleatoric (homoscedastic) task-uncertainty \cite{Kendall}. Our analyses shows that uncertainty loss weights result in the least NT behavior across two orthogonal datasets (image recognition and pharmacogenomics). In settings with harder tasks and smaller sample sizes such as pharmacogenomics, we envisage that our approach will enable robust MTL and reduce NT cases.
\section{Methods}
Our approach combines and expands upon ideas from Deep-AMTFL \cite{Lee2} and Kendall \cite{Kendall}. We expand Deep-AMTFL as follows: Let $\mathbf{f}^{\mathbf{w_t}}(\mathbf{x})$ be Deep-AMTFL's output function for input $\mathbf{x}$ and task-weight $\mathbf{w_t}$. Following Kendall \cite{Kendall}, we aim to obtain $\sigma_t$, the task observation noise parameter (\textit{aleatoric homoscedastic uncertainty}). In a classification setting, we obtain $\sigma_t$ as follows: \begin{equation}p\left(\mathbf{y_t} \mid \mathbf{f}^{\mathbf{w_t}}(\mathbf{x}), \sigma_t\right)=\operatorname{Softmax}\left(\frac{1}{\sigma_t^{2}} \mathbf{f}^{\mathbf{w_t}}(\mathbf{x})\right)
\end{equation} and sample from the resulting probability vector. Similarly, in a regression setting, $\sigma_t$ can be obtained as follows:
\begin{equation}p\left(\mathbf{y_t} \mid \mathbf{f}^{\mathbf{w_t}}(\mathbf{x}), \sigma_t\right)=\frac{1}{2 \sigma_{t}^{2}} \|\mathbf{y}_{t}-\mathbf{f}^{\mathbf{w_t}}(\mathbf{x})\|^{2}
+\log \sigma_{t}\end{equation} We extend $\sigma_t$ to \textit{T} different tasks $t \in\{1, \ldots, T\}$ to obtain the relative task-uncertainties in a MTL setting. We allow $\sigma$ to be learnable with the network parameters $\boldsymbol{W}$. We call this approach U-AMTFL.

\section{Experiments}
\subsection{MNIST - Imbalanced }This setting follows \cite{Lee2}, where an imbalanced version of the MNIST dataset is sampled. Sample numbers decrease from the digit zero to nine, making higher digits harder to predict.
\subsubsection{Baselines}
Following \cite{Lee2}, we have: 1) CNN. A baseline CNN model (LeNet like). 2) MTL-CNN, the extended CNN in an MTL manner with a shared layer L1-regularized. 3) Deep-AMTFL (CNN) 4) U-AMTFL (CNN). All models follow the hyperparameters suggested in \cite{Lee2}. \subsubsection{Evaluation and Results}
\begin{figure}
\centering
\subfloat[Average classification error]{\label{a}\includegraphics[width=.5\linewidth]{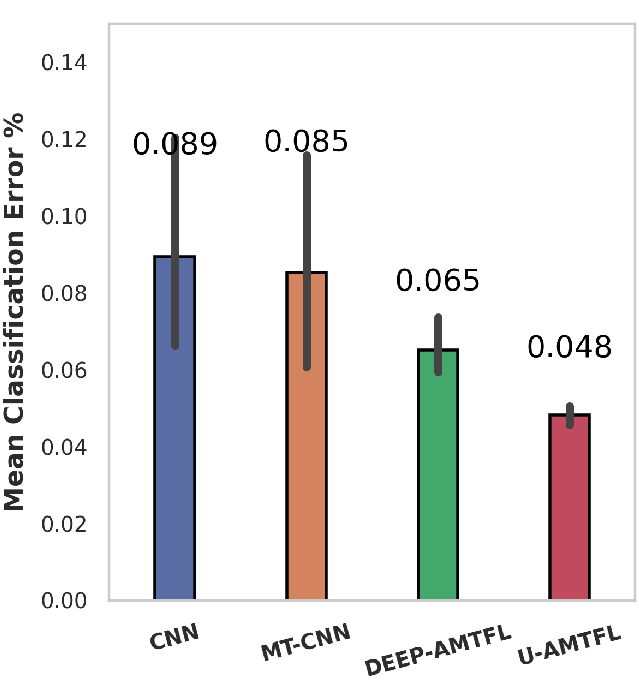}}\hfill
\subfloat[Per task improvements]{\label{b}\includegraphics[width=.5\linewidth]{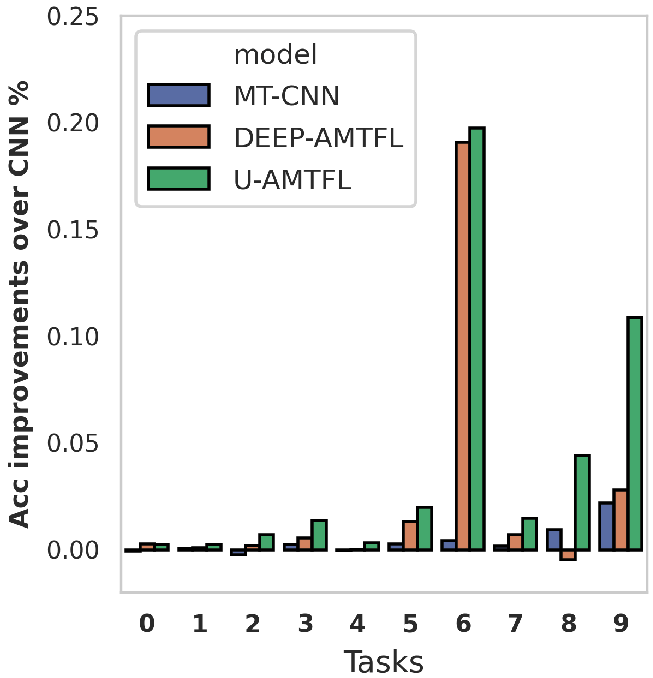}}
\caption{MNIST Results}
\label{fig1}
\end{figure}

We report the average accuracy error over five runs of baselines in Figure \ref{fig1}a. U-AMTFL outperformed all other baselines. In particular, we investigate per task improvement over the CNN. We observed that U-AMTFL presents a large improvement in later tasks, 8 and 9 (Figure \ref{fig1}b), tasks with small sample sizes. U-AMTFL is the only model not facing NT and presenting improved accuracy on all tasks. This result shows the effectiveness of using task-uncertainty as a reliability proxy in asymmetric knowledge transfer.
\subsection{Pharmacogenomics (GDSC)}
Here, we evaluate MTL models on a pharmacogenomics dataset. GDSC \cite{Iorio2016} is a dataset of drug screening responses on panels of cancer cell lines. Along with drug responses, GDSC provides gene expression data. Predicting cancer drug response given a transcriptomic profile is still bounded by the scarcity of samples. Also, only a few cell lines are responsive to a given drug. We aim to evaluate how NT is presented in this setting.
\subsubsection{Baselines}
1) STL. We use the linear regression proposed by GDSC \cite{Iorio2016}. 2) MTL-NN, Deep-AMTFL, and U-AMTFL, present the same architecture as in the previous section, but with the CNN replaced with a linear layer followed by ReLU. We keep the same hyperparameters. This setup is a regression problem and we aim to predict the IC50 (minimal concentration that kills 50\% of cells) values for 181 drugs at the same time, using cancer cell line gene expression data as features. We corrected 805 cell line gene expression data for library-size and log-transformed it, followed by mean-centering standardization.
\subsubsection{Evaluation and Results}
\begin{figure}
\centering
\subfloat[Mean error improvement over STL]{\label{a}\includegraphics[width=.5\linewidth]{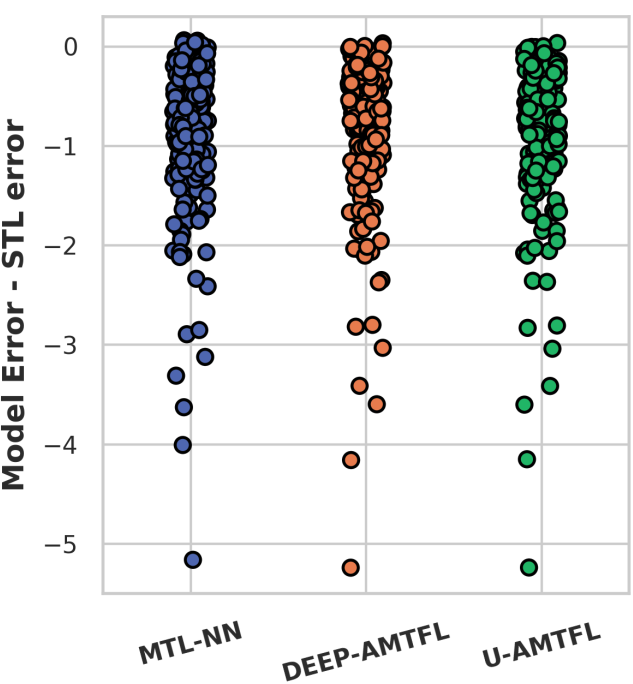}}\hfill
\subfloat[Negative Transfer Tasks]{\label{b}\includegraphics[width=.5\linewidth]{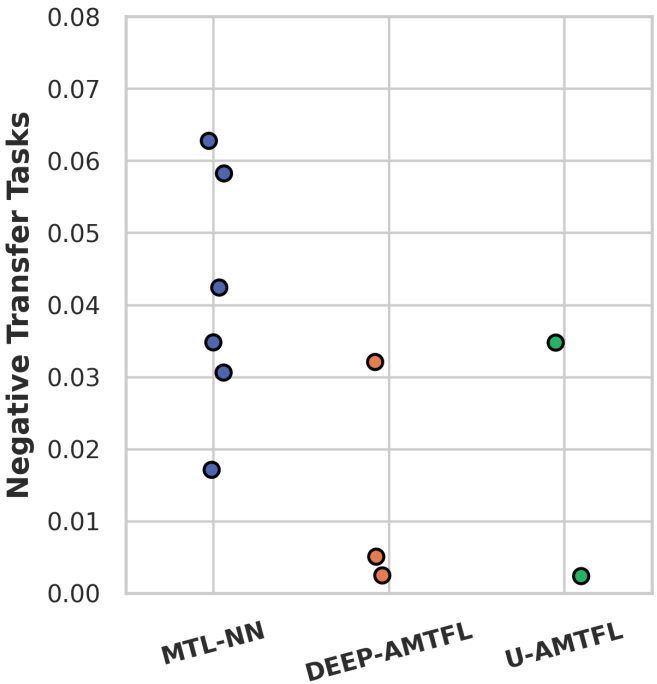}}
\caption{Pharmacogenomics Results}
\label{fig2}
\end{figure}

We report the mean MSE subtracted from the mean STL MSE over three runs in Figure \ref{fig2}a. Positive values indicate NT. We observed that all MTL methods performed to a large extent better than STL. The averaged error performance is STL with 2.6210, MTL-NN with 1.7436, Deep-AMTFL with 1.7374 and U-AMTFL with 1.7342. Prominently, we report positive errors in Figure \ref{fig2}(b), to dissect the distribution of NT cases. U-AMTFL outperformed all baselines, with the fewest number of NT cases, only two. Being able to improve performance with close to zero NT cases shows the effectiveness of uncertainty weighted asymmetric knowledge transfer.

\section{Conclusion}
We proposed to reduce negative transfer (NT) in asymmetric multi-task learning by using uncertainty (\textit{aleatoric homoscedastic}) as a proxy for task-reliability. We presented improved results in two different domains. Based on the present results, we aim to further investigate NT in pharmacogenomics, especially when predicting response for drugs with fewer samples and by exploring feature uncertainty and its relationship to different tasks.

\bibliography{aaai21.bib}

\begin{thebibliography}{4}
\providecommand{\natexlab}[1]{#1}
\providecommand{\url}[1]{\texttt{#1}}
\providecommand{\urlprefix}{URL }
\expandafter\ifx\csname urlstyle\endcsname\relax
  \providecommand{\doi}[1]{doi:\discretionary{}{}{}#1}\else
  \providecommand{\doi}{doi:\discretionary{}{}{}\begingroup
  \urlstyle{rm}\Url}\fi

\bibitem[{{Iorio et al.}(2016)}]{Iorio2016}
{Iorio et al.} 2016.
\newblock A Landscape of Pharmacogenomic Interactions in Cancer.
\newblock \emph{Cell} 166(3): 740--754.
\newblock ISSN 0092-8674.

\bibitem[{{Kendall et al.}(2018)}]{Kendall}
{Kendall et al.} 2018.
\newblock Multi-Task Learning Using Uncertainty to Weigh Losses for Scene
  Geometry and Semantics.
\newblock In \emph{2018 IEEE/CVF Conference on Computer Vision and Pattern
  Recognition}, 7482--7491.

\bibitem[{{Lee et al}(2016)}]{Lee1}
{Lee et al}. 2016.
\newblock Asymmetric Multi-Task Learning Based on Task Relatedness and Loss.
\newblock ICML'16.

\bibitem[{{Lee et al.}(2018)}]{Lee2}
{Lee et al.} 2018.
\newblock Deep Asymmetric Multi-task Feature Learning.
\newblock volume~80 of \emph{Proceedings of Machine Learning Research},
  2956--2964. PMLR.

\end{thebibliography}

\end{document}